\documentclass[conference]{IEEEtran}
\ifCLASSINFOpdf
\else
\fi
\usepackage{graphicx}
\usepackage{subfig, endnotes, }
\usepackage{float}
\usepackage{hyperref}
\usepackage{multirow}
\usepackage{tabularx}
\usepackage[table]{xcolor}
\usepackage{url}
\usepackage{pgfplotstable}
\usepackage{amssymb,latexsym,amsmath}
\usepackage[ruled,vlined]{algorithm2e}

\makeatletter
\renewcommand{\@algocf@capt@plain}{above}
\newcommand*{\rom}[1]{\expandafter\@slowromancap\romannumeral #1@}
\makeatother
\begin{document}
%
\title{Triclustering of Gene Expression Microarray data using Evolutionary Approach}


\author{\IEEEauthorblockN{Shreya Mishra}
\IEEEauthorblockA{\small Dept. of Computer Sc. Engineering\\
IIIT Bhubaneswar, 751003\\
Email: shreya.mishra693@gmail.com}
\and
\IEEEauthorblockN{Swati Vipsita}
\IEEEauthorblockA{\small Dept. of Computer Sc. Engineering\\
IIIT Bhubaneswar, 751003\\
Email: swati@iiit-bh.ac.in}}


%


\maketitle

\begin{abstract}
In Tri-clustering, a sub-matrix is being created, which exhibit highly similar behavior with respect to genes, conditions and time-points. In this technique, genes with same expression values are discovered across some fragment of time points, under certain conditions. In this paper, triclustering using evolutionary algorithm is implemented using a new fitness function consisting of 3D Mean Square residue (MSR) and Least Square approximation (LSL). The primary objective is to find triclusters with minimum overlapping, low MSR, low LSL and covering almost every element of expression matrix, thus minimizing the overall fitness value. To improve the results of algorithm, new fitness function is introduced to find good quality triclusters. It is observed from experiments that, triclustering using EA yielded good quality triclusters. The experiment was implemented on yeast Saccharomyces dataset.\\
Index Terms-Tri-clustering, Genetic Algorithm, Mean squared residue, Volume, Weights, Least square approximation.
\end{abstract}


\section{INTRODUCTION}
Technological research has been revolutionized using high power processing techniques and has highly increased chunks of data available. Specifically, gene expression data have transformed research in biological field by its potential to supervise RNA concentration change in large number of genes concurrently available \cite{brown1999exploring}.

Clustering techniques are applied for analysis of gene expression matrix which creates subset of genes which display same expression behavior \cite{rubio2008classification}. Conventional clustering algorithms are applied on complete dataset, testing each gene under all conditions. Clustering algorithms does not always give best results because most of the gene patterns exist under a subset of experimental conditions. Hence, clustering mechanism should be shifted to methods which can discover local patterns from gene expression data.

Bi-clustering is able to find local patterns by finding gene patterns under a subset of experimental conditions. Still, both clustering and bi-clustering are not enough when gene expression micro array data is analyzed and focus is drawn on effect of time on gene's behavior. These type of longitudinal experiments allow thorough analysis of molecular processes where time plays a crucial role. Cell cycles, evolution of diseases and development at molecular level are some of the examples \cite{bar2004analyzing}. So, specific tools must be used for analysis of genes which are analyzed under specific conditions and time points. Hence, tri-clustering emerges as an effective tool for the same.

Tri-clustering algorithms find genes with similar expression in a subset of conditions along a time segment. Coherent tri-cluster is defined as set of genes that expresses same numerical value in certain time and condition or same behavior despite of exactly same numerical value i.e positively or negatively correlated changes in expression values of genes. Both positive and negative coherent clusters and regulation reaction among genes are useful for identifying effective phenotypes.

Triclustering using evolutionary computation (genetic algorithm) is presented in this paper which searches for group of genes that express similar patterns across both condition and time points i.e in three dimensional space. Most of the clustering and biclustering approaches define similarity of elements on the basis of distance measures \cite{tibshirani1999clustering} but these measures are not accurate to find similarities in genes as correlations might still exist in some genes inspite of having different magnitude levels. Thus, two different evaluation measures are combined: one being mean square residue (MSR) and other being least square approximation (LSL). MSR is used for finding coherent triclusters, which is a 3-D form of MSR derived from biclustering distance measuring technique used for discovering similar gene patterns. LSL is least square approximation (LSL) which finds triclusters  with coherent behavior by calculating distance between slopes of least square lines of the tricluster discovered.

In 2005, Zhao and Zaki \cite{zhao2005tricluster} proposed algorithm named triCluster which extracts three dimensional pattern from gene expression data. Set of measures are also defined to calculate goodness of triclusters. In 2006, \cite{jiang2006gtricluster}, a generalized form of triclustering (g-triCluster) was defined which focused on finding more coherent triclusters which are resistant to noise. In \cite{braga2008partricluster}, the authors labeled the triclustering problem in association with its NP-completeness. They proposed a filter-labeled stream with a parallel approach, hence, showing greater improvement in computational cost. Later in \cite{yin2007mining}, coherent triclusters found regulatory relationships among genes. This algorithm was applied on both real and synthetic datasets. In \cite{wang2010efficiently}, an algorithm was proposed to find time delayed clusters. It discovered cycle time of gene expression which is necessary for forming a network of gene regulation.

In \cite{xu2009finding}, a new approach named LagMiner was introduced which finds time-lagged three dimensional clusters. This algorithm was able to remove constraint of coherence, size of subspace, number of genes, regulation and length of time period. All the above algorithms were implemented on real and synthetic dataset. Tri-clustering is also implemented using evolutionary computation. Specifically, the multi-objective algorithm optimizes the most conflicting requirements of a tricluster. This algorithm was implemented on real data set \cite{liu2008multi}.

In \cite{sim2010discovering}, a strategy to find triclusters in real valued data was introduced. Major concern in discovering tricluster was to find subspaces with large number of elements from gene expression matrix but this should not be the only concern in triclustering approach. So, an approach was proposed \cite{hu2010algorithm} which aimed to find triclusters with low variance and could mine quantitative data. This algorithm was tested for synthetic data set and on cross-species genome data set, thus being a major advancement in this topic.

A triclustering approach to find temporal dependent association rules in micro array data was introduced in \cite{liu2010novel}. The rules generated were able to present regulated relations among genes.

Remaining paper is arranged as : section \rom{2} discusses the basic concepts of tri-clustering, section \rom{3} explains the EA based algorithm implemented to derive triclusters. The experiment details, numerical simulation results are detailed in section \rom{4}. Section \rom{5} concludes the paper.

\section{PARADIGM OF TRICLUSTERING}
Tri-clustering appear as an expansion of bi-clustering which focuses on evolution of genes at different time points under certain experimental conditions. So, from a gene expression matrix EM, which consist of genes $G_{EM}$, conditions $C_{EM}$, and time points $T_{EM}$, tri-clustering is defined as a procedure that finds triclusters $T_{1},…,T_{n}$ from EM, where a tricluster T is defined as T =G x C x T, where $G \subset G_{EM}$, $C \subset C_{EM}$, and $T \subset T_{EM}$, ie, a subset of genes from gene expression matrix EM that carries information about the behavior of some genes under conditions C at times T.

Two quality measures are defined below which are crucial to get good quality triclusters. These definitions are defined in \cite{gutierrez2014trigen} .
\begin{itemize}
\item[(a)] Three Dimensional Mean Squared Residue:- This definition is a modification of MSR used in bicluster quality measurement to three dimensional MSR which evaluates the similarity of triclusters which contain subsets of genes, conditions, and time points. This measure is denoted as $MSR_{3D}$.
	\begin{equation}
	MSR_{3D}=\frac{\sum\limits_{g\in G,c\in C,t\in T} R^{2}_{gct}}{G_{1}*C_{1}*T_{1}}
	\end{equation}
	where $R_{gct}$ is measured as\\
	\begin{equation}
	\begin{aligned}
	R_{gct}=TC_{v}(t,g,c)+M_{GC}(t)+M_{GT}(c)+M_{CT}(g)- \\
	M_{G}(c,t)-M_{C}(g,t)-M_{T}(g,c)-M_{GCT}
	\end{aligned} 
	\end{equation}
where\\
$M_{GC}(t)$ : Mean of all gene values under conditions at a time t \\
$M_{GT}(c)$ : Mean of all gene values across time in certain condition c \\
$M_{CT}(g)$ : Mean of gene g in time under conditions \\
$M_{G}(c,t)$ : Mean of all gene values in a particular condition and time \\
$M_{C}(g,t)$ : Mean of values of gene at time under conditions \\
$M_{T}(g,c)$ : Mean of a gene under a condition in all times\\
$M_{GCT}$ : Mean of all values in tricluster.\\
	
A graphical depiction of terms used in eq. (1) can be understood from Fig. 1. It can be seen that $MSR_{3D}$ calculates similarity of the tricluster on the basis of difference of each element of gene expression $TC_{V}(i, j, k)$ (Fig. 1(a)), mean of all conditions at all times for a gene g $M_{CT}(g)$ (Fig. 1(b)), mean of all genes at all times for a condition c $M_{GT}(c)$ (Fig. 1(c)), mean of all genes under all conditions at time t $M_{GC}(t)$ (Fig. 1(d)) with the mean of a condition c and a time t under all genes $M_{G}(c, t)$ (Fig. 1(e)), mean of a gene g and a time t under all conditions $M_{C}(g, t)$ (Fig. 1(f)), mean of a gene g and a condition c under all times $M_{T}(g, c)$ (Fig. 1(g)), and the mean of all values in TC $M_{GCT}$ (Fig. 1(h)). If the value of $MSR_{3D}$ is closer to zero, then the tricluster is more homogeneous. $MSR_{3D}$ is also able to find negatively correlated genes.
	\begin{figure}[!hbpt]
		\centering
		\includegraphics[height=10cm,width=\linewidth]{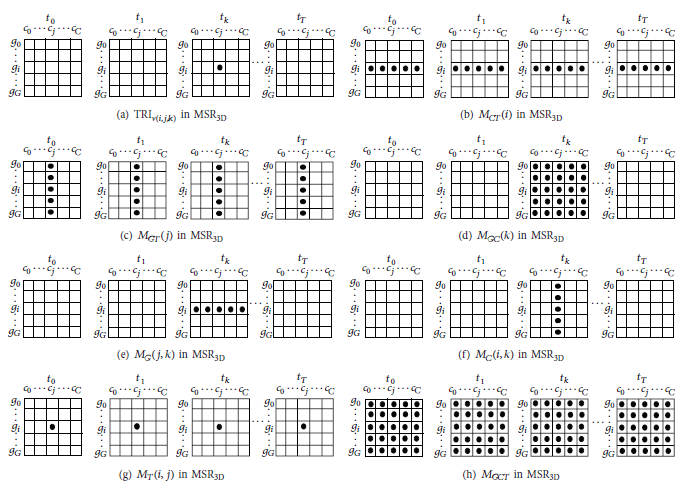}
		\caption{$MSR_{3D}$ structural members}
	\end{figure}

\item[(b)] Least Square Approximation:- This term is defined by equation:
	\begin{equation}
	LSL=\frac{T_{r}+C_{r}+G_{r}}{3}
	\end{equation}
It measures similarity of the least squares approximation of points in each graphic of the three views which depicts a tricluster. Firstly, for every time coordinate, conditions are represented on x-axis, expression values on y-axis and genes are outlined ($T_{r}$ in (3)). Secondly, for every condition coordinate, times  on x-axis, expression values on y-axis and genes are outlined ($C_{r}$ in (3)). Thirdly, for every condition coordinate, genes  on x-axis, expression values on y-axis and times are outlined ($G_{r}$ in (3)). A representation of this can be seen in Fig. 2. All elements of numerator in Eq. (3) have values indicated in equation group (4, 5) in common.
	\begin{figure}
	\includegraphics[height=4cm,width=\linewidth]{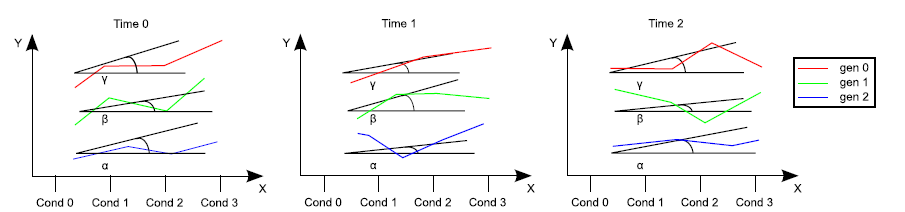}
	\caption{A solution with 3 genes, 4 conditions, 3 time is shown. This figure illustrate a view for every time point, condition in x-axis, expression values in y-axis, genes are then outlined. Slope of every least square approximation $\alpha,~ \beta  ~and ~ \gamma$ are compared to find correlation between patterns. Similar step is followed for rest of the views also.}
\end{figure}
	\begin{align}
	sumX_{tc}=\sum_{g\in G}g \qquad \qquad sumXX_{tc}=\sum_{g\in G}g^{2} 
	\end{align}
	\begin{align}
	sumX_{g}=\sum_{t\in T}t \qquad \qquad sumXX_{g}=\sum_{t\in T}t^{2} 
	\end{align}
where $sumX_{tc}$ summarizes all genes of the individuals undergoing evaluation, $sumXX_{tc}$ gives squared summation of all genes, $sumX_{g}$ gives summation of all times and $sumXX_{g}$ gives squared summation of all times. In first view, distance among all least square approximations are calculated from Eq. (6) ($T_{r}$).
	\begin{equation}
	T_{r} = \frac{\sum_{t_i, t_j\in T}|TD_{t_{i}}-TD_{t_{j}}|}{(T_{l}-1)*T_{l}}
	\end{equation}
	\begin{equation}
	\forall t\in T,\qquad TD_t=\frac{G_l*sumXY_{t}-(sumX_{tc}*sumY_{t})}{G_l*sumXX_{tc}-sumX^2_{tc}}
	\end{equation}
	\begin{equation}
	\forall t\in T,\qquad sumXY_{t}=\sum_{g\in G}\sum_{c\in C} g*IN_{v}(t, g, c)
	\end{equation}
	\begin{equation}
	\forall t\in T,\qquad sumY_{t}=\sum_{g\in G}\sum_{c\in C} IN_{v}(t, g, c)
	\end{equation}
where TD in (7) represent measures, $sumXY_{t}$ and $sumY_{t}$ are sum of each expression value of each time point, every combination of particular gene and conditions multiplied by genes in (8) and for every particular time, sum of all expression values at that time and every combination of particular gene and condition as mentioned in (9).
	
Similarly, $C_{r}$ is defined by Eq. (10) which gives distance between every least square approximation generated in second view.
	\begin{equation}
	C_{r} = \frac{\sum_{c_i, c_j\in C}|CD_{c_{i}}-CD_{c_{j}}|}{(C_{l}-1)*C_{l}}
	\end{equation}
	\begin{equation}
	\forall c\in C,\qquad CD_c=\frac{G_l*sumXY_{c}-(sumX_{tc}*sumY_{c})}{G_l*sumXX_{tc}-sumX^2_{tc}}
	\end{equation}
	\begin{equation}
	\forall c\in C,\qquad sumXY_{c}=\sum_{t\in T}\sum_{g\in G} g*IN_{v}(t, g, c)
	\end{equation}
	\begin{equation}
	\forall c\in C,\qquad sumY_{c}=\sum_{t\in T}\sum_{g\in G} IN_{v}(t, g, c)
	\end{equation}
where CD in  (11) represents measures, $sumXY_{c}$ and $sumY_c$ gives values for every particular condition, sum of every expression value of that condition, every combination of particular time and gene multiplied by genes in (12) and for every particular condition, sum of all expression values at that paticular condition and every combination of particular time and gene as mentioned in (13).
	
Eq. (14) gives $G_{r}$ which calculate distance between every least square approximation generated in third view.
	\begin{equation}
	G_{r} = \frac{\sum_{c_i, c_j\in C}|GD_{c_{i}}-GD_{c_{j}}|}{(C_{l}-1)*C_{l}}
	\end{equation}
	\begin{equation}
	\forall c\in C,\qquad GD_c=\frac{T_l*sumXY_{g}-(sumX_{g}*sumY_{g})}{T_l*sumXX_{g}-sumX^2_{g}}
	\end{equation}
	\begin{equation}
	\forall c\in C,\qquad sumXY_{c}=\sum_{t\in T}\sum_{g\in G} t*IN_{v}(t, g, c)
	\end{equation}
	\begin{equation}
	\forall c\in C,\qquad sumY_{c}=\sum_{t\in T}\sum_{g\in G} IN_{v}(t, g, c)
	\end{equation}
where GD represents measures (14) and $sumXY_{c}$ and $sumY_{c}$ are for every particular condition, sum of every expression value of that condition, every combination of particular gene and time multiplied by time (15) and for every particular condition, sum of all expression values at that particular condition and every combination of particular gene and time as mentioned in (16).
\end{itemize}

\section{EVOLUTIONARY ALGORITHM FOR TRI-CLUSTERING}
TriEA is implemented on the basis of genetic algorithm. Following steps are involved in this evolutionary process: initial population is generated in initialization step which minimizes overlapping with already discovered triclusters, another is evaluation step where quality of each individual is calculated using a fitness function; a selection step decides which individuals should be allowed to survive to next generation, crossover exchanges genetic material between pairs of individuals to create offspring and mutation flips a particular individual bit to make sure genetic variability is maintained of future generations.

\subsection{CHROMOSOME ENCODING}
A potential tricluster solution TC is represented by each individual of population. It contains genetic information which is operated upon by the genetic operators. An individual chromosome consist of three sequences: one series of genes G, conditions C, and time points T. Fig. 3 illustrates the chromosome generation of triclusters.
\begin{center}
	\begin{figure}
		\caption{Genetic algorithm codification}
		\centering
		\includegraphics[height=4cm,width=5cm]{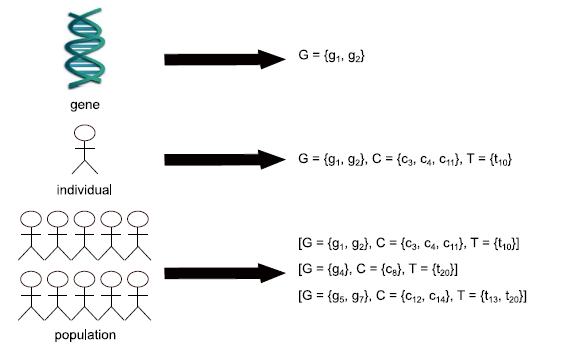}
	\end{figure}
\end{center}
 These chromosomes are formed on the basis of expression matrix,
\begin{equation}
G = <g_{i_{1}} , g_{i_{2}} ,…, g_{i_{A}} >
\end{equation}
where A is number of genes in the expression data, $i_{j}<i_{j+1}$ for all genes and $1<i_{j}<A$.
Similarly: 
\begin{equation}
C = <c_{i_{1}} , c_{i_{2}} ,…, c_{i_{B}} >
\end{equation}
where B is number of conditions in expression data, $i_{j}<i_{j+1}$ for all genes, and $1<i_{j}<B$.
Finally, T represents different time stamps:
\begin{equation}
T = <t_{i_{1}} , t_{i_{2}} ,…, t_{i_{C}} >
\end{equation}
where C is number of gene samples measured over time,$ t_{i_{1}} < t_{i_{2}} <,…,< t_{i_{C}}$ .

So, the chromosome bit string in the population is formed in following sequence
\begin{equation}
\begin{aligned}
Chromosome= <g_{i_{1}} , g_{i_{2}} ,…, g_{i_{A}} > |  <c_{i_{1}} , c_{i_{2}} ,…,\\
c_{i_{B}} > | <t_{i_{1}} , t_{i_{2}} ,…, t_{i_{C}} >
\end{aligned}
\end{equation}

Triclusters are represented in the form of binary strings of X+Y+Z length, X being genes, Y being conditions and Z being time points in expression matrix. If value of bit in individual is 1, it indicates respective gene, condition  or time point have a place in that tricluster.

Let binary string be of 15 bits (5 for genes, 5 for conditions and 5 for time points). The genotypic representation of chromosome will be:
\begin{center}
	10110$|$10001$|$11001
\end{center} 

\subsection{INITIAL POPULATION GENERATION}
The population is initialized randomly. A subset of genes, conditions and times are randomly generated, and tricluster TC = T x C x G are assigned. Remaining individuals are also randomly generated but keeping into consideration that individuals should be non overlapping with already generated individuals. In order to achieve this, every time a new individual is initialized, its indices of genes, conditions and times are stored. After that new individuals are initialized from random subset that did not appear in any previous individual.

\subsection{OPERATION OF CROSSOVER}
Parent $P_{1}$ and $P_{2}$ individuals are combined to create two offsprings $O_{1}$ and $O_{2}$. A probability of crossover ($P_{c}$) is associated with this operation. Genetic material of parents is combined by a random crosspoint in the chromosome and mixing the coordinates to form offsprings.

Formally, let $P_{1}^{i}$ and $P_{2}^{i}$ be two parents at some iteration i. Resultant offsprings are
\begin{equation}
(O_1^i , O_2^i)=f_{C, G, T} (P_1^i, P_2^i)
\end{equation}
where, f randomly selects subset of (C, G, T) from parents with crossover probability ($p_{c}^{i}$).

The procedure can be formalized as follows. Let $S_{1}$ = $<C_{i}^{1} ,…, C_{i}^{m} >$ and $S_{2}$ = $<C^{'}_{j_{1}} ,…, C^{'}_{j_{n}}>$ be two sequences of chromosome, where $C_{i_{k}} , C_{j_{l}} \in  i_{1} <...< i_{m},~ j_{1} < .. < j_{n}$ and m, n $<$ length of chromosome.

Let P be number randomly generated between 1 and min of (m, n). Two new offsprings are formed as follows:
\begin{equation}
O_{1}=<C_{i_{1}}, ..., C_{i_{p}}, C_{j_{p+1}}, ... ,C_{j_{n}}>
\end{equation}
\begin{equation}
O_{2}=<C^{'}_{j_{1}}, ... ,C^{'}_{j_{1}}, C^{'}_{i_{p+1}}, ... , C^{'}_{i_{m}}>
\end{equation}

\subsection{OPERATION OF MUTATION}
Depending on the probability of mutation ($P_m$), any chromosome can be mutated. Standard mutation is applied which flips the value of a single bit randomly in chromosome. 

\subsection{SELECTION OPERATOR}
In the reproduction stage, offsprings are created by making them compete with other individuals for the placement in next generation. Elitism is applied which means only the best chromosomes should be allowed to survive to next generation. Tournament selection operator is applied with size of 2.

\subsection{PROPOSED FITNESS EQUATION}
Fitness of every individual helps algorithm to find best solutions which are carried forward in further generations. Two quality measures are combined in TriEA algorithm, one being 3D Mean Square Residue measure (MSR), referred as $f_{MSR}$. The second one is $f_{LSL}$.

\textbf{Weights term}
\begin{equation}
Weights=G_l*w_g+C_l*w_c+T_l*w_t
\end{equation}
where $w_g$, $w_c$ and $w_t$ represents weights for genes, conditions and times in the tricluster. Higher value of weights indicate that triclusters with many components are found by triEA.\\
\textbf{Distinction term}
\begin{equation}
Distinction=\frac{CDN_g}{G_l}*wd_g+\frac{CDN_c}{C_l}*wd_c+\frac{CDN_t}{T_l}*wd_t
\end{equation}
where $CDN_g$ (Coordinate Distinction Number of g) : Genes which are not present in tricluster solution being evaluated. \\
$CDN_c$ (Coordinate Distinction Number of c) : Conditions which are not present in tricluster solution being evaluated. \\
$CDN_t$ (Coordinate Distinction Number of t) : Time points which are absent in tricluster solution being evaluated.\\
$wd_g$, $wd_c$, $wd_t$ are weights assigned at prior to genes, conditions and time points. Distinction calculates difference between chromosome under evaluation and triclusters already discovered.

If the value of weights is increased, non-overlapping tricluster solutions can be discovered.\\
\textbf{Fitness Equation}\\
Following equation defines this function:
\begin{equation}
F(TC)=MSR+LSL-Weights-Distinction
\end{equation}
This is a minimization equation which consist of four factors , Weight facto Distinction factor, particular MSR, LSL factors.

\begin{algorithm}[]
	\caption{TRICLUSTERING USING EA }
	\SetAlgoLined
	\KwResult{Tricluster solutions with LSL less than given threshold}
	Load EM\\
	\While{maxiter$<$=no\textunderscore of\textunderscore tricluster}{
		Initialize Population\\
		Evaluate Population\\
		\While{iter$<$=no\textunderscore of\textunderscore generation}{
			Selection of Individuals\\
			Crossover among individuals\\
			Mutation of Individuals\\
			Evaluation  of Individuals\\
		}
		Best\textunderscore Individual=Individual with least fitness\\
		\If{LSL(Best\textunderscore Individual)$<$ $\delta$}{
			Store in final results
		} 
		return final results	
	} 
\end{algorithm}

\section{NUMERICAL SIMULATION RESULTS AND DISCUSSION}
The algorithm TriEA is implemented using R i386 3.3.2 and executed on system of 32 bit operating system and Intel(R)Core(TM)i7-4790 CPU @ 360GHz processor with 4 GB of RAM. TriEA is applied to the yeast (Saccharomyces cerevisiae) cell data \cite{spellman1998comprehensive}. The objective of this project is to find genes whose mRNA levels are controlled by cell cycle processes. When TriEA is applied to dataset of such form, meaningful sequences in cell cycle can be discovered. In this experiment, 6179 genes are examined under 6 conditions named clb2, cln3, cdc15, cdc28, pheromone and elutriation. Gene profiles are captured at 2 time points for cln3, 18 for pheromone, 2 for clb2, 24 for cdc15, 17 for cdc28 and 14 for elutriation. For applying TriEA, conditions with only 2 time points are not considered and first 14 time points of pheromone, elutriation, cdc15, cdc28 are considered in dataset. Hence, the expression matrix consist of 6179 genes, 4 conditions and 14 time points. 200 genes are taken to carry out this experiment due to its increased search space and high computation cost. The original expression data is preprocessed using max-min normalization method such that all the values of matrix lies between 0 to l.
\begin{equation}
em(i,j)'=\frac{em(i,j)-col_{min}}{col_{max}-col_{min}}
\end{equation}

Parameter values for triEA are given in TABLE \rom{1}. Values that are missing in dataset are replaced with randomly generated values between 0 and 1.
\begin{table}[h]
	\caption{Parameter settings for TriEA algorithm}
	\centering
	\begin{tabular}{|l|l|}
		\hline
		Parameter Name           & TriEA\\ \hline
		Population Size          & 20 	\\ \hline
		Number of generation    & 100   \\ \hline
		Probability of Crossover  & 0.95 \\ \hline
		Probability of Mutation   & 0.50 \\ \hline
		Weight for conditions        & 0.1  \\ \hline
		Weight for genes             & 0.1  \\ \hline
		Weight for time             & 0.1   \\ \hline
		Genes from original dataset taken    & 200  \\ \hline
		Threshold $\delta$            & 1050  \\ \hline
		Number of triclusters    & 20  \\ \hline
	\end{tabular}
\end{table} 

\begin{table}[!htbp]
	\caption{Information about triclusters found by TriEA}
	\begin{tabular}{|p{0.50cm}|l|l|p{0.6cm}|l|l|}
		\hline
		Tric No & Fitness Value  & LSL & Weight & Distinction & MSR \\ \hline
		1  &    6246.74 &  19.74 & 1.0  & 0.0505 &   6228.04 \\ \hline
		2  &  139141.44 & 492.01 & 7.7 & 0.0043 & 138657.13 \\ \hline
		3  &    429.41 &  12.59 & 0.8 & 0.0280 &    417.64 \\ \hline
		4  &   4003.45 & 662.39 & 1.7 & 0.0106 &   3342.77 \\ \hline
		5  &   4120.64 & 576.88 & 1.6 & 0.0113 &   3545.37 \\\hline
		6  &     432.86 &  55.89 & 0.8 & 0.0280 &    377.79 \\\hline
		7  &    2837.64 & 123.84 & 1.0 & 0.0218 &   2714.82 \\\hline
		8  &   10885.81 & 819.77  & 2.5 & 0.0077 &  10068.54 \\\hline
		9  &   10763.06 & 834.15 & 1.6 & 0.0109 &   9930.51 \\\hline
		10 &    4533.08 & 745.21 & 1.6 & 0.0113 &   3789.48 \\\hline
		11 &   11241.41 & 171.32  & 1.4 & 0.0129 &  11071.50 \\\hline
		12 &    7714.68 & 352.65 & 2.2 & 0.0085 &   7364.23 \\\hline
		13 &   14278.17 & 693.94 & 1.8 & 0.0095 &  13586.042 \\\hline
		14 &   16013.35 & 134.89  & 1.2 & 0.0209 &  15879.68 \\\hline
		15 &   25446.38 & 125.77 & 3.4 & 0.0063 &  25324.015 \\\hline
		16 &   55523.39 & 737.98 & 4.9 & 0.0052 &  54790.31 \\\hline
		17 &    6966.46 &  49.03 & 2.1 & 0.0088 &   6919.50 \\\hline
		18 &   14581.88 & 225.15 & 2.5 & 0.0077 &  14359.23 \\\hline
		19 &    3589.76 & 255.23  & 1.5 & 0.0120 &   3336.03 \\\hline
		20 &   23074.08 & 873.73 & 1.9 & 0.0089 &  22202.25  \\\hline
	\end{tabular}
\end{table}

\begin{figure}[!htbp]
	\caption{GA Convergence for triEA algorithm for 100 generations}
	\centering
	\includegraphics[height=6cm,width=8cm]{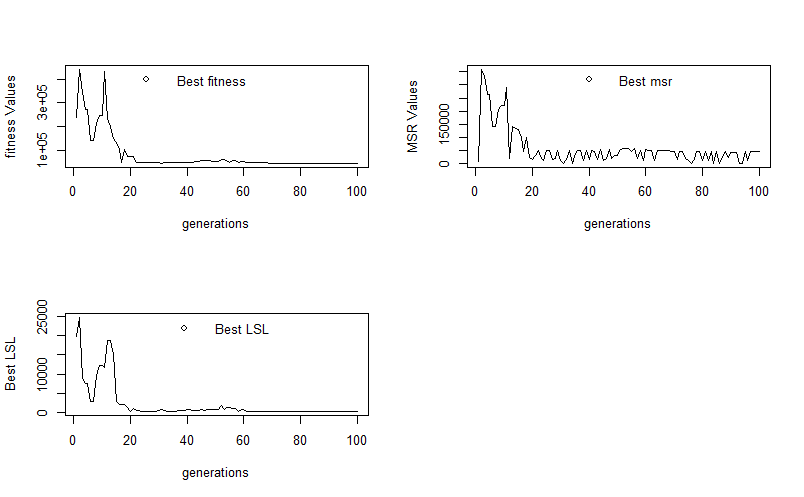}
\end{figure}
After the convergence of TriEA, 20 resultant triclusters were derived with minimum MSR and LSL score and it was observed all of them have least square approximation (LSL) score less than the threshold $\delta$ i.e 1050. The detailed results are shown in TABLE \rom{2}. The average value of LSL for all tri-clusters obtained is 398.12. Tricluster obtained with least LSL value is 12.59. Also, average value of MSR for all triclusters obtained is 17545.24 which is quite large because this algorithm focus on triclusters with minimum LSL value but not least MSR value of triclusters. Least MSR value obtained is 417.64. Due to large MSR values of triclusters, overall fitness value also increases.

From Fig. 4 convergence of GA for 100 generations shows that triclusters with minimum MSR and LSL are preferred in evolution process resulting in to minimum fitness value of the tricluster. Thus, in order to find triclusters with minimum overall fitness value, triclusters with minimum MSR and LSL should be preferred in evolution process.

The results in TABLE \rom{2} are of TriEA for 200 genes, 4 conditions and 14 time points.

Fig. 4 shows convergence of GA towards minimum value for 100 generations.

\section{CONCLUSION AND FUTURE WORK}
For Tri-clustering, TriEA is implemented for 200 genes using two quality measures i.e MSR and LSL. It can be concluded from results that GA converges better when both the quality measures are combined to form one fitness function. TriEA yields good quality triclusters with least overlapping among triclusters discovered. But the tricluster quality can be improved more, if both MSR and LSL are minimized by algorithm simultaneously.  

The microarray data is mined for longitudinal experiments using algorithm TriEA, but this algorithm can also be applied in other biological domains, for example gene expression data can be combined with gene regulatory data by methods of replacing dimensions of time by chip-chip information depicting transcription factor gene interactions that provides information about  gene regulatory networks. This algorithm can be applied on mining RNA-seq data repositories.

\bibliographystyle{IEEEtran}
\bibliography{Bibliography}

\begin{thebibliography}{10}
\providecommand{\url}[1]{#1}
\csname url@samestyle\endcsname
\providecommand{\newblock}{\relax}
\providecommand{\bibinfo}[2]{#2}
\providecommand{\BIBentrySTDinterwordspacing}{\spaceskip=0pt\relax}
\providecommand{\BIBentryALTinterwordstretchfactor}{4}
\providecommand{\BIBentryALTinterwordspacing}{\spaceskip=\fontdimen2\font plus
\BIBentryALTinterwordstretchfactor\fontdimen3\font minus
  \fontdimen4\font\relax}
\providecommand{\BIBforeignlanguage}[2]{{%
\expandafter\ifx\csname l@#1\endcsname\relax
\typeout{** WARNING: IEEEtran.bst: No hyphenation pattern has been}%
\typeout{** loaded for the language `#1'. Using the pattern for}%
\typeout{** the default language instead.}%
\else
\language=\csname l@#1\endcsname
\fi
#2}}
\providecommand{\BIBdecl}{\relax}
\BIBdecl

\bibitem{brown1999exploring}
P.~O. Brown and D.~Botstein, ``Exploring the new world of the genome with dna
  microarrays,'' \emph{Nature genetics}, vol.~21, pp. 33--37, 1999.

\bibitem{rubio2008classification}
C.~Rubio-Escudero, F.~Mart{\'\i}nez-{\'A}lvarez, R.~Romero-Zaliz, and I.~Zwir,
  ``Classification of gene expression profiles: comparison of k-means and
  expectation maximization algorithms,'' in \emph{Hybrid Intelligent Systems,
  2008. HIS'08. Eighth International Conference on}.\hskip 1em plus 0.5em minus
  0.4em\relax IEEE, 2008, pp. 831--836.

\bibitem{bar2004analyzing}
Z.~Bar-Joseph, ``Analyzing time series gene expression data,''
  \emph{Bioinformatics}, vol.~20, no.~16, pp. 2493--2503, 2004.

\bibitem{tibshirani1999clustering}
R.~Tibshirani, T.~Hastie, M.~Eisen, D.~Ross, D.~Botstein, P.~Brown
  \emph{et~al.}, ``Clustering methods for the analysis of dna microarray
  data,'' \emph{Dept. Statist., Stanford Univ., Stanford, CA, Tech. Rep}, 1999.

\bibitem{zhao2005tricluster}
L.~Zhao and M.~J. Zaki, ``Tricluster: an effective algorithm for mining
  coherent clusters in 3d microarray data,'' in \emph{Proceedings of the 2005
  ACM SIGMOD international conference on Management of data}.\hskip 1em plus
  0.5em minus 0.4em\relax ACM, 2005, pp. 694--705.

\bibitem{jiang2006gtricluster}
H.~Jiang, S.~Zhou, J.~Guan, and Y.~Zheng, ``gtricluster: a more general and
  effective 3d clustering algorithm for gene-sample-time microarray data,'' in
  \emph{International Workshop on Data Mining for Biomedical
  Applications}.\hskip 1em plus 0.5em minus 0.4em\relax Springer, 2006, pp.
  48--59.

\bibitem{braga2008partricluster}
R.~Braga~Ara{\'u}jo, G.~H. Trielli~Ferreira, G.~H. Orair, W.~Meira, R.~A.
  Celso~Ferreira, D.~Olavo Guedes~Neto, and M.~J. Zaki, ``The partricluster
  algorithm for gene expression analysis,'' \emph{International Journal of
  Parallel Programming}, vol.~36, no.~2, pp. 226--249, 2008.

\bibitem{yin2007mining}
Y.~Yin, Y.~Zhao, B.~Zhang, and G.~Wang, ``Mining time-shifting co-regulation
  patterns from gene expression data,'' in \emph{Advances in data and web
  management}.\hskip 1em plus 0.5em minus 0.4em\relax Springer, 2007, pp.
  62--73.

\bibitem{wang2010efficiently}
G.~Wang, L.~Yin, Y.~Zhao, and K.~Mao, ``Efficiently mining time-delayed gene
  expression patterns,'' \emph{IEEE Transactions on Systems, Man, and
  Cybernetics, Part B (Cybernetics)}, vol.~40, no.~2, pp. 400--411, 2010.

\bibitem{xu2009finding}
X.~Xu, Y.~Lu, K.-L. Tan, and A.~K. Tung, ``Finding time-lagged 3d clusters,''
  in \emph{Data Engineering, 2009. ICDE'09. IEEE 25th International Conference
  on}.\hskip 1em plus 0.5em minus 0.4em\relax IEEE, 2009, pp. 445--456.

\bibitem{liu2008multi}
J.~Liu, Z.~Li, X.~Hu, and Y.~Chen, ``Multi-objective evolutionary algorithm for
  mining 3d clusters in gene-sample-time microarray data,'' in \emph{Granular
  Computing, 2008. GrC 2008. IEEE International Conference on}.\hskip 1em plus
  0.5em minus 0.4em\relax IEEE, 2008, pp. 442--447.

\bibitem{sim2010discovering}
K.~Sim, Z.~Aung, and V.~Gopalkrishnan, ``Discovering correlated subspace
  clusters in 3d continuous-valued data,'' in \emph{Data Mining (ICDM), 2010
  IEEE 10th International Conference on}.\hskip 1em plus 0.5em minus
  0.4em\relax IEEE, 2010, pp. 471--480.

\bibitem{hu2010algorithm}
Z.~Hu and R.~Bhatnagar, ``Algorithm for discovering low-variance 3-clusters
  from real-valued datasets,'' in \emph{Data Mining (ICDM), 2010 IEEE 10th
  International Conference on}.\hskip 1em plus 0.5em minus 0.4em\relax IEEE,
  2010, pp. 236--245.

\bibitem{liu2010novel}
Y.-C. Liu, C.-H. Lee, W.-C. Chen, J.~Shin, H.-H. Hsu, and V.~S. Tseng, ``A
  novel method for mining temporally dependent association rules in
  three-dimensional microarray datasets,'' in \emph{Computer Symposium (ICS),
  2010 International}.\hskip 1em plus 0.5em minus 0.4em\relax IEEE, 2010, pp.
  759--764.

\bibitem{gutierrez2014trigen}
D.~Guti{\'e}rrez-Avil{\'e}s, C.~Rubio-Escudero, F.~Mart{\'\i}nez-{\'A}lvarez,
  and J.~C. Riquelme, ``Trigen: A genetic algorithm to mine triclusters in
  temporal gene expression data,'' \emph{Neurocomputing}, vol. 132, pp. 42--53,
  2014.

\bibitem{spellman1998comprehensive}
P.~T. Spellman, G.~Sherlock, M.~Q. Zhang, V.~R. Iyer, K.~Anders, M.~B. Eisen,
  P.~O. Brown, D.~Botstein, and B.~Futcher, ``Comprehensive identification of
  cell cycle--regulated genes of the yeast saccharomyces cerevisiae by
  microarray hybridization,'' \emph{Molecular biology of the cell}, vol.~9,
  no.~12, pp. 3273--3297, 1998.

\end{thebibliography}
\cleardoublepage
%
\IEEEpeerreviewmaketitle
\end{document}